# Learning to Coordinate Efficiently:
# A Model-based Approach

**Ronen I. Brafman**                                    BRAFMAN@CS.BGU.AC.IL
*Computer Science Department*
*Ben-Gurion University*
*Beer-Sheva, Israel 84105*

**Moshe Tennenholtz**                                   MOSHE@ROBOTICS.STANFORD.EDU
*Faculty of Industrial Engineering and Management*
*Technion*
*Haifa 32000, Israel*

## Abstract

In common-interest stochastic games all players receive an identical payoff. Players participating in such games must learn to coordinate with each other in order to receive the highest-possible value. A number of reinforcement learning algorithms have been proposed for this problem, and some have been shown to converge to good solutions in the limit. In this paper we show that using very simple model-based algorithms, much better (i.e., polynomial) convergence rates can be attained. Moreover, our model-based algorithms are guaranteed to converge to the optimal value, unlike many of the existing algorithms.

## 1. Introduction

In some learning contexts, the learning system is actually a collection of components, or agents, that have some common goal or a common utility function. The distributed nature of such systems makes the problem of learning to act in an unknown environment more difficult because the agents must coordinate both their learning process and their action choices. However, the need to coordinate is not restricted to distributed agents, as it arises naturally among self-interested agents in certain environments. A good model for such environments is that of a *common-interest stochastic game* (CISG). A *stochastic game* (Shapley, 1953) is a model of multi-agent interactions consisting of multiple finite or infinite stages, in each of which the agents play a one-shot strategic form game. The identity of each stage depends stochastically on the previous stage and the actions performed by the agents in that stage. The goal of each agent is to maximize some function of its reward stream - either its average reward or its sum of discounted rewards. A CISG is a stochastic game in which at each point the payoff of all agents is identical.

Various algorithms for learning in CISGs have been proposed in the literature. These include specialized algorithms for certain classes of CISGs, (e.g., Claus & Boutilier, 1997; Wang & Sandholm, 2002), as well as more general algorithms for learning in general stochastic games that converge to a Nash equilibrium (e.g., Littman, 1994; Hu & Wellman, 1998; Littman, 2001; Bowling & Veloso, 2001) and can be applied to CISGs. For many of these, results that show convergence in the limit to an equilibrium exist. However, all current algorithms suffer from at least one of the following when applied to CISGs: (1) When multiple equilibria exist, there is no guarantee of convergence to an optimal equilibrium. (2) They are not known to be efficient – i.e., at best, they have convergence in the limit guarantees. In addition, many of the algorithms make strong assumptions about the information obtained by the agents during the course of the game. In particular, the agents know what actions are available to each other in each stage of the game, and they can observe the choices made by the other agents. This latter assumption is called *perfect monitoring* in the game-theory literature. This is to be contrasted with *imperfect monitoring* in which it is assumed that each agent can only observe its own payoff, but it cannot observe the actions performed by the other players.

In this paper we show that using a simple model-based approach, we can provide algorithms with far better theoretical guarantees, and often, in more general settings than previous algorithms. More precisely, we provide a learning algorithm with polynomial-time convergence to a near-optimal value in stochastic common-interest games under imperfect monitoring. In the case of repeated games, our algorithm converges to the actual optimal value. The key to our result is a recent powerful learning algorithm, R-MAX (Brafman & Tennenholtz, 2002). In particular, the following aspect of R-MAX plays a crucial role: it is a *deterministic* learning algorithm that guarantees polynomial-time convergence to near-optimal value in the single-agent case.[1] We note that in our treatment of stochastic games we make the standard assumption that the state is fully observable. That is, the agents recognize the "situation" in which they are in at each point in time.

In the following section, we provide the necessary background on stochastic games. In Section 3, we discuss the basic algorithm and its particular variants. In Section 4 we conclude the paper with a discussion of some important variants: $n$-player common-interest stochastic games and repeated games. Because an understanding of R-MAX and its convergence properties is needed for a more complete picture of our results, we provide this information in the Appendix.

## 2. Background

We start with a set of standard definitions and a discussion of the basic assumptions of our approach.

### 2.1 Game Definitions

A *game* is a model of multi-agent interaction. In a *strategic form game*, we have a set of players, each of whom chooses some action to perform from a given set of actions. As a result of the players' combined choices, some outcome is obtained which is described numerically in the form of a payoff vector, i.e., a vector of values, one for each of the players. We concentrate on two-player games. General $n$-person games can be treated similarly.

A common description of a two-player strategic-form game is in the form of a matrix. The rows of the matrix correspond to player 1's actions and the columns correspond to player 2's actions. The entry in row $i$ and column $j$ in the game matrix contains the rewards obtained by the players if player 1 plays his $i^{th}$ action and player 2 plays his $j^{th}$ action. The set of possible combined choices of the agents are called *joint actions*. Thus, if player 1 has $k$ possible actions and player 2 has $l$ possible actions, there are $k \cdot l$ joint actions in the game. Without loss of generality, we assume that the payoff of a player in a game is taken from the interval of real numbers $P = [0, R_{max}]$.

---

1. In fact, R-MAX converges in zero-sum stochastic games as well, but we will not need this capability.

A *common-interest* game is a game in which the rewards obtained by all agents are identical. A *coordination* game is a common-interest game in which all agents have the same set $A$ of actions, and, in addition, for every $i$, the reward all agents receive for selecting the joint action $(i, i)$ is greater than the reward they receive for selecting $(i, j)$ for any $j \neq i$.

In a *repeated game* the players play a given game $G$ repeatedly. We can view a repeated game, with respect to a game $G$, as consisting of an infinite number of iterations, for each of which the players have to select an action of the game $G$. After playing each iteration, the players receive the appropriate payoffs, as dictated by that game's matrix, and move to the next iteration.

A *stochastic game* is a game in which the players play a possibly infinite sequence of one-shot, strategic form games from some given set of games, which we shall take to be finite. After playing each game, the players receive the appropriate payoff, as dictated by that game's matrix, and move to a new game. The identity of this new game depends, stochastically, on the previous game and on the players' actions in that previous game. Formally:

**Definition 1** *A two player, stochastic-game $M$ on states $S = \{1, \ldots, N\}$, and actions $A = \{a_1, \ldots, a_k\}$, consists of:*

- **Stage Games:** *each state $s \in S$ is associated with a two-player game in strategic form, where the action set of each player is $A$.*

- **Probabilistic Transition Function:** $tr(s, t, a, a')$ *is the probability of a transition from state $s$ to state $t$ given that the first player plays $a$ and the second player plays $a'$.*

A stochastic game is similar to a Markov decision process (MDP). In both models actions lead to transitions between states of the world. The main difference is that in an MDP the transition depends on the action of a single player whereas in a stochastic game the transition depends on the joint action of both players. In addition, in a stochastic game, the reward obtained by the player for performing an action depends on its action *and* the action of the other player. To model this, we associate a game with every state. Therefore, we often use the terms *state*, *stage game*, and *game* interchangeably.

Much like in regular one-shot games, we can restrict our attention to special classes of stochastic and repeated games. In particular, in this paper we are interested in common-interest stochastic games (CISGs), i.e., stochastic games in which each stage-game is a common-interest game. CISGs can be viewed as a special extension of MDPs. They correspond to an MDP in which the agent is actually a distributed system. That is, it has different components, all of which influence the choice of action, but all of which work toward a common goal. Thus, for every CISG, we define its *induced MDP* to be the MDP obtained from the CISG when we assume the existence of a single agent that controls all the agents in the CISG. That is, a single agent whose set of actions corresponds to the *joint*-actions of the CISG agents. Because the payoffs for all agents are identical, this is a well-defined concept.

Each history of length $t$ in a game with perfect monitoring consists of a sequence of $t$ tuples of the form (stage-game, joint actions, payoffs), followed by the last state reached. Thus, the set of possible histories of length $t$ in a game with perfect monitoring is $(S \times A^2 \times P)^t \times S$. We denote the set of histories of any length in this setting by $H_p$. Similarly, in the case of imperfect monitoring, each history is a sequence of tuples of the form (stage-game, single-agent action, single-agent payoff), followed by the last state reached. Thus, the set of possible histories of length $t$ in a game with imperfect monitoring is $(S \times A \times P)^t \times S$, and the set of possible histories in this setting is denoted $H_{imp}$.

Given a CISG with imperfect monitoring (respectively, perfect monitoring), a *policy* for the agent is a mapping from $H_{imp}$ (respectively, $H_p$) to the set of possible probability distributions over $A$. Hence, a policy determines the probability of choosing each particular action for each possible history. A *pure* policy is a deterministic policy. A *stationary* policy is one in which the action is a function of the last state only.

In defining the value of a CISG, we treat it much like an MDP because the ultimate goal is to provide a coordinated strategy for the agents that will lead to the maximal reward. We use the *average expected reward criterion*. Given a CISG $M$ and a natural number $T$, $U(s, \pi, \rho, T)$ denotes the expected $T$-step undiscounted average reward when player 1 follows policy $\pi$ and player 2 follows policy $\rho$ starting from state $s$. The optimal $T$-step value for $M$ starting at $s$ is $U(s, T) = \max_{(\pi, \rho)} U(s, \pi, \rho, T)$. We define $U(s) = \liminf_{T \to \infty} U(s, T)$, and refer to it as the *value* of $s$

## 2.2 Assumptions, Complexity and Optimality

The setting we consider is one in which agents have a common interest. We assume that these agents are aware of the fact that they have a common interest and that they know the number of agents that participate in this game. This is most natural in a distributed system. Moreover, we assume that they all employ the same learning algorithm; this latter assumption is common to most work in coordination learning. It is quite natural if the agents are part of a distributed system, or if there exists a standard, commonly accepted algorithm for coordination learning.

We make a number of additional assumptions: First, we assume that the agent always recognizes the identity of the stage-game it reached, but not its associated payoffs and transition probabilities. Second, we assume that the maximal possible reward, $R_{max}$ is known ahead of time. Finally, we restrict our attention to ergodic CISGs, which, in a sense, are the only class of CISGs for which the type of result we seek is possible. Kearns and Singh introduced the *ergodicity* assumption in the context of their $E^3$ algorithm for learning in MDPs (Kearns & Singh, 1998). A similar assumption was used by Hoffman and Karp (1966) in the context of stochastic games, where it is referred to as *irreducibility*. An MDP is said to be *ergodic* if the Markov-chain obtained by fixing any pure stationary policy is ergodic. That is, if any state is reachable from any other state. We restrict our attention to CISGs whose induced MDP is ergodic.

Irreducible stochastic games and ergodic MDPs have a number of nice properties, as shown by Hoffman and Karp (1966). First, the maximal long-term average reward is independent of the starting state, implying that $U(s)$ is actually independent of $s$. Thus, from now on we use v(M) to denote the *optimal value* of the game $M$, which is identical to $U(s)$ for all states $s$. Second, this optimal value can be obtained by a stationary policy, i.e., by a policy that depends on the current stage-game only. In particular, in the case of CISGs, this optimal value can be obtained using a pure stationary strategy.

In a sense, there seems to be little point in theoretical discussions of learning in the context of MDPs and CISGs that are not ergodic. It is evident that in order to learn the agent must perform some sort of implicit or explicit exploration. If the MDP is not ergodic, the agent's initial choices can have irreversible consequences on its long-term value. But one cannot expect the agent to always make correct initial choices when it has no knowledge of the game.

Next, we wish to discuss the central parameter in the analysis of the complexity – the *mixing time* – first identified by Kearns and Singh (1998). Kearns and Singh argue that it is unreasonable to refer to the efficiency of learning algorithms without referring to the efficiency of convergence to a desired value given *complete information*. They defined the *ε-return mixing time* of a policy $\pi$ (in an MDP) to be the smallest value of $T$ such that for all $t > T$, and for any starting state $s$, the expected average reward of $\pi$ is at least $\epsilon$ close to the expected infinite horizon average reward of $\pi$. That is, letting $U(s, \pi, t)$ denote the expected $t$-step average reward of $\pi$ starting at $s$, define $U(\pi) = \liminf_{t \to \infty} U(s, \pi, t)$, where this latter value does not depend on the choice of $s$ because of the ergodicity assumption. The $\epsilon$-return mixing time of $\pi$, $T$ satisfies:

$$T = \min_{T' \geq 1} \forall s, t' \geq T' : U(s, \pi, t') > U(\pi) - \epsilon.$$

Notice that the mixing time has nothing to do with learning. It is influenced by the stochastic nature of the model and its interaction with the policy – however obtained. We adopt their definition, but with respect to the MDP induced by the CISG of interest.

Finally, we define a *near-optimal polynomial-time learning* algorithm for CISGs to be a learning algorithm $\mathcal{A}$ such that given a CISG $M$, some $\epsilon > 0, \gamma > 0$ and $0 < \delta < 1$, there exists some $T > 0$ that is polynomial in $1/\epsilon, 1/\delta, 1/\gamma$, the $\epsilon$-return mixing time of an optimal policy, and the size of the description of the game, such that for every $t > T$ the actual $t$-step average reward obtained by $\mathcal{A}$ is at least $(1 - \gamma)(v(M) - 2\epsilon)$, with a probability of at least $1 - \delta$.

Notice that the above definition requires that if we consider any particular, fixed $t > T$, then the actual average reward will be close to optimal with an overwhelming probability.

## 3. The Algorithms

The following is our basic result:

**Theorem 1** *There exists an near-optimal polynomial time algorithm for learning in two-player CISGs under imperfect monitoring provided that all agents have a shared polynomial bound* [2] *on the number of actions of each other.*

We note that in certain special sub-cases, e.g., under perfect monitoring, there will be no need for the multiplicative factor of $1 - \gamma$.

---

2. Formally, we will assume that there is a constant $k \geq 1$, such that if the number of actions available to agent $i$ is $f_i$, then it is commonly known that any agent can have a most of $f = (max(f_1, f_2))^k$ actions. Other ways of formalizing this concept are treated similarly.

An immediate corollary of Theorem 1 is:

**Corollary 1** *There exists a near-optimal polynomial time algorithm for learning in CISGs under imperfect monitoring provided that all agents know that all stage games are coordination games.*

**Proof:** Coordination games are a special instance of CISGs. In particular, in these games we have an identical number of actions for all players, so each player has a polynomial bound on the number of actions available to the other players. □

The proof of Theorem 1 will proceed in stages using successively weaker assumptions. The reduction will be from a particular multi-agent case to the case of single-agent learning in MDPs. The settings in the first cases is quite natural for a distributed system, whereas the latter cases make quite weak assumptions on the agents' knowledge of each other. To succeed in our proof, the single-agent learning algorithm to which the problem is reduced must have the following properties:

- It is a near-optimal polynomial time algorithm. That is, given $\epsilon > 0$ and $0 < \delta < 1$, then with a probability of $1 - \delta$, the algorithm, when applied to an MDP, leads to an actual average payoff that is $\epsilon$-close to the optimal payoff for any $t > T$, where $T$ is polynomial in $1/\epsilon, 1/\delta$, the $\epsilon$-return mixing time, and the description of the MDP.

- The precise value of $T$ can be computed ahead of time given $\epsilon$, $\delta$, the $\epsilon$-return mixing time of the optimal policy, and the description of the MDP.

- The algorithm is deterministic. That is, given a particular history, the algorithm will always select the same action.

One such algorithm is R-MAX (Brafman & Tennenholtz, 2002), described in detail in the Appendix. In order to make the presentation more concrete, and since, at the moment, R-MAX appears to be the only algorithm with these properties, we explicitly refer to it in the proof. However, any other algorithm with the above properties can be used instead.

We now proceed with the description of the reductions. Note: In Cases 1-5 below, we assume that the $\epsilon$-return mixing time of an optimal policy, denoted by $T_{mix}$ is known. For example, in repeated games, $T_{mix}$ is 1. In stochastic games, on the other hand, $T_{mix}$ is unlikely to be known. However, this assumption is useful for later development. In Case 6, we remove this assumption using a standard technique. Finally, in all cases we assume that the learning accuracy parameters – $\epsilon$, $\delta$, and $\gamma$ – are known to all agents.

**Case 1: Common order over joint actions.** This case is handled by reduction to the single-agent case.

If all agents share an ordering over the set of joint actions in every stage game, then they can emulate the single-agent R-MAX algorithm on the induced MDP. This follows from the fact that R-MAX is deterministic. Thus, the agents start with an initially identical model. They all know which joint action a single-agent would execute initially, and so they execute their part of that joint action. Since all agents observe their own reward always. And because this is identical to all agents, the update of the model will be the same for all agents. Thus, the model of all agents is always identical. Therefore, in essence we have a distributed, coordinated execution of the R-MAX algorithm, and the result follows immediately from the properties of R-MAX. Note that Case 1 does not require perfect monitoring.

We note that many cases can be reduced to this case. Indeed, below we present techniques for handling a number of more difficult cases in which an ordering does not exist, that work by reduction to Case 1. We note that in special games in which there is sufficient asymmetry in the game (e.g., different action set sizes, different off-diagonal payoffs) it is possible to devise more efficient methods for generating an ordering (e.g., via randomization).

**Case 2: A commonly known ordering over the agents, and knowledge of the action-set size of all agents.** First, suppose that the agents not only know the action set sizes of each other, but also share a common ordering over the actions of each agent. This induces a natural lexicographic ordering over the joint actions based on the ordering of each agent's actions and the ordering over the agents. Thus, we are back in case 1.

In fact, we do not really need a common ordering over the action set of each agent. To see this, consider the following algorithm:

- Each agent will represent each joint action as a vector. The $k^{th}$ component of this vector is the index of the action performed by agent $k$ in this joint action. Thus, $(3, 4)$ denotes the joint action in which the first agent according to the common ordering performs its third action, and the second agent performs its fourth action.

- At each point, the agent will choose the appropriate joint action using R-MAX. The choice will be identical for all agents, provided the model is identical.

- Because each agent only performs its aspect of the joint action, there cannot be any confusion over the identities of the actions. Thus, if R-MAX determines that the current action is $(3, 4)$, then agent 1 will perform its third action and agent 2 will perform its fourth action.

The model is initialized in an identical manner by all agents. It is updated identically by all agents because at each point, they all choose the same vector and their rewards are identical. The fact that when the chosen joint action is $(3, 4)$ agent 1 does not know what precisely is the fourth action of agent 2 does not matter.

Following Case 2, we can show that we can deal efficiently with a perfect monitoring setup.

**Case 3: Perfect Monitoring.** We can reduce the case of perfect monitoring to Case 2 above as follows. First, we make the action set sizes known to all agents as follows: In the first phases of the game, each agent will play its actions one after the other, returning to its first action, eventually. This will expose the number of actions. Next, we establish a common ordering over the agents as follows: Each agent randomly selects an action. The agent that selected an action with a lower index will be the first agent. Thus, we are now back in Case 2.

We note that much of the work on learning coordination to date has dealt with case 3 above in the more restricted setting of a repeated game.

**Case 4: Known action set sizes.** What prevents us from using the approach of Case 2 is the lack of common ordering over the agents. We can deal with this problem as follows: In an initial order exploration phase, each agent randomly selects an ordering over the agents and acts based on this ordering. This is repeated for $m$ trials. The length of each trial is $T'$, where $T'$ is such that R-MAX will return an actual average reward that is $\epsilon$-close to optimal after $T'$ steps with a probability of at least $1 - \delta/2$, provided that an identical ordering was chosen by all agents. $m$ is chosen so that with high probability in one of the trials all agents will choose the same ordering.

Once the order exploration phase is done, each agent selects the ordering that led to the best average reward in the order exploration phase and uses the policy generated for that ordering for $Q$ steps. $Q$ must be large enough to ensure that the losses during the order exploration phase are compensated for – an explicit form is given below. Hopefully, the chosen ordering is indeed identical. However, it may be that it was not identical but simply led to a good average reward in the order exploration phase. In the exploitation phase, it could however lead to lower average rewards. Therefore, while exploiting this policy, the agent checks after each step that the actual average reward is similar to the average reward obtained by this policy in the order exploration phase. $T'$ is chosen to be large enough to ensure that a policy with $\epsilon$-return mixing time of $T_{mix}$ will, with high probability yield an actual reward that is $\epsilon$-close to optimal. If the average is not as expected, then with high probability the chosen ordering is not identical for all agents. Once this happens, the agents move to the next best ordering. Notice that such a change can take place at most $m - 1$ times.

Notice that whenever we realize in step $T$ that the actual average reward is too low, a loss of at most $\frac{R_{max}}{T}$ is generated. By choosing $T'$ such that $\frac{R_{max}}{T'} < \frac{\epsilon}{m}$, we get that we remain in the scope of the desired payoff (i.e., $2\epsilon$ from optimal). The $1/m$ factor is needed because this error can happen at most $m$ times.

We now show that $Q$ and $m$ can be chosen so that the running time of this algorithm is still polynomial in the problem parameters, and that the actual average reward will be $\epsilon$-close to optimal times $1 - \gamma$ with probability of at least $1 - \delta$. First, notice that the algorithm can fail either in the order exploration phase or in the exploitation phases. We will ensure that either failure occurs with a probability of less the $\delta/2$.

Consider the order-exploration phase. We must choose $m$ so that with probability of at least $1 - \delta/2$ an iteration occurs in which:

- An identical ordering is generated by all agents;
- R-MAX leads to an $\epsilon$-close to optimal average reward.

In a two-player game, the probability of an identical ordering being generated by both agents is $1/2$. With an appropriate, and still polynomial, choice of $T'$, the probability of R-MAX succeeding given an identical ordering is at least $1 - \delta/2$. We now choose $m$ so that $[1 - (1/2 \cdot (1 - \delta/2))]^m < \delta/2$, and this can be satisfied by an $m$ that is polynomial in $\delta$. Notice that this bounds the failure probability in selecting the right ordering and obtaining the desired payoff, and that the above inequality holds whenever $m > \frac{log(\delta/2)}{log([1-(1/2 \cdot (1-\delta/2))])} > |log(\delta/2)|$, which is polynomial in the problem parameters.

Next, consider the exploitation phases. Given that we succeeded in the exploration phase, at some point in the exploitation phase we will execute an $\epsilon$-optimal policy for $Q$ iterations. If $Q$ is sufficiently large, we can ensure that this will yield an actual average reward that is $\epsilon$-close to optimal with a probability that is larger than $1 - \delta/2$. Thus, each failure mode has less than $\delta/2$ probability, and overall, the failure probability is less than $\delta$.

Finally, we need to consider what is required to ensure that the average reward of this whole algorithm will be as high as we wanted. We know that during the $Q$ exploration steps an $\epsilon$-close to optimal value is attained. However, there can be up to $m-1$ repetitions of $T$-step iterations in which we obtain a lower expected reward. Thus, we must have that $\frac{Q \cdot (v(M)-\epsilon)}{Q+2mT} > (1-\gamma) \cdot (v(M) - \epsilon)$. This is satisfied if $Q > \frac{1}{\gamma}2mT$ which is still polynomial in the problem parameters.

**Case 5: Shared polynomial bound on action-set sizes.** We repeat the technique used in Case 4, but for every chosen ordering over the agents, the agents systematically search the space of possible action-set sizes. That is, let $Sizes$ be some arbitrary total ordering over the set of all possible action set sizes consistent with the action-set size bounds. The actual choice of $Sizes$ is fixed and part of the agents' algorithm. Now, for each action sets size in $Sizes$, the agents repeat the exploration phase of Case 4, as if these were the true action sets sizes. Again, the agents stick to the action-set size and ordering of agents that led to the best expected payoff, and they will switch to the next best ordering and sizes if the average reward is lower than expected.

Let $b$ be the agents' bound on their action sets sizes. The only difference between our case and Case 4 is that the initial exploration phase must be longer by a factor of $b^2$, and that in the exploitation phase we will have to switch $O(m \cdot b^2)$ policies at most. In particular, we now need to attempt possible orderings for the $b^2$ possibilities for the size of the joint actions set. Thus, the failure probabilities are treated as above, and $Q$ needs to be increased so that $\frac{Q \cdot (v(M)-\epsilon)}{Q+2mb^2T} > (1-\gamma) \cdot (v(M) - \epsilon)$. $Q$ remains polynomial in the problem parameters.

**Case 6: Polynomial bound on action set sizes, unknown mixing time.** In this case we reach the setting discussed in Theorem 1 in its full generality. To handle it, we use the following fact: R-MAX provides a concrete bound on the number of steps required to obtain near-optimal average reward for a given mixing time. Using this bound, we can provide a concrete bound for the running time of all the cases above, and in particular, Case 5. This enables us to use the standard technique for overcoming the problem of unknown mixing time that was used in $E^3$ (Kearns & Singh, 1998) and in R-MAX. Namely, the agents run the algorithm for increasing values of the mixing time. For every fixed mixing time, we operate as in Case 4–5, but ensuring that we achieve a multiplicative factor of $1 - \gamma/2$, instead of $1-\gamma$. Suppose that the correct mixing time is $T$. Thus, when the agents run the algorithm for an assumed mixing time $t$ smaller than $T$, we expect an average reward lower than optimal, and we need to compensate for these losses when we get to $T$. These losses are bounded by the optimal value times the number of steps we acted sub-optimally. This number of steps is polynomial in the problem parameters. Suppose that when $T$ is reached, we are suddenly told that we have reached the true mixing time. At that point, we can try to compensate for our previous losses by continuing to act optimally for some period of time. The period of time required to compensate for our previous losses is polynomial in the problem parameters.

Of course, no one tells us that we have reached the true mixing time. However, whenever we act based on the assumption that the mixing time is greater than $T$, but still polynomial in $T$, our actual average reward will remain near optimal. Thus, the stages of the algorithm in which the assumed mixing time is greater than $T$ automatically compensate for the stages in which the mixing time is lower than $T$. Thus, there exists some $T'$ polynomial in $T$ and the other problem parameters, such that by the time we run Case 5 under the assumption that $T'$ is the mixing time, we will achieve an $\epsilon$-close to optimal reward with a multiplicative factor of $1 - \gamma/2$. In addition, when $\frac{T}{T'} < (1 - \gamma/2)$ we get that the multiplicative loss due what happened before we have reached the mixing time $T$ is another $(1 - \gamma/2)$ factor. Altogether, when we reach the assumed mixing time of $T'$, we get the desired value with a multiplicative factor of $1 - \gamma$.

Having shown how to handle Case 6, we are done with the proof of Theorem 1.  □

## 4. Discussion

We showed that polynomial-time coordination is possible under the very general assumption of imperfect monitoring and a shared polynomial bound on the size of the action sets. To achieve this result, we relied on the existence of a deterministic algorithm for learning near-optimal policies for MDPs in polynomial time. One such algorithm is R-MAX, and it may be possible to generate a deterministic variant of $E^3$ (Kearns & Singh, 1998).

Our algorithms range from a very simple and extremely efficient approach to learning in perfect monitoring games, to conceptually simple but computationally more complex algorithms for the most general case covered. Whether these latter algorithms are efficient in practice relative to algorithms based on, e.g., Q-learning, and whether they can form a basis for more practical algorithms, remains an interesting issue for future work which we plan to pursue. Regardless of this question, our algorithms demonstrate clearly that coordination-learning algorithms should strive for more than convergence in the limit to some equilibrium, and that more general classes of CISGs with a more restrictive information structure can be dealt with.

In Section 3, we restricted our attention to two-player games. However, we used a more general language that applies to $n$-player games so that the algorithms themselves extend naturally. The only issue is the effect on the running time of moving from a constant number of agents to $n$-player games. First, observe that the size of the game matrices is exponential in the number of agents. This implies that little changes in the analysis of the earlier cases in which a shared fixed ordering over the set of agents exists. The complexity of these cases increases exponentially in the number of agents, but so does the size of the game so the algorithm remains polynomial in the input parameters.

If we do not have a shared ordering over the agents, then we need to consider the set of $n!$ possible orderings. Notice however that although the number of orderings of the players is exponential in the number of players, using the technique of case 4, we get that with overwhelming probability the players will choose identical orderings after time that is polynomial in the number of such orderings. More specifically, Chernoff bound implies that after $P(n!)$ trials, for a polynomial $P$ implied by the bound, we will get with overwhelming probability that the same ordering on actions will be selected by all agents. Hence, although this number might be large (when $n$ is large), if the number of agents, $n$, is smaller than the number of actions available, $a$, this results in a number of iterations that is polynomial in the representation of the game, since the representation of game requires $a^n$ entries. Naturally, in these cases it will be of interest to consider succinct representations of the stochastic game, but this goes beyond the scope of this paper.

One special class of CISGs is that of common-interest repeated games. This class can be handled in a much simpler manner because there is no need to learn transition functions and to coordinate the transitions. For example, in the most general case (i.e., imperfect monitoring) the following extremely simple, yet efficient algorithm can be used: the agents play randomly for some $T$ steps, following which they select the first action to provide the best immediate reward. If $T$ is chosen appropriately, we are guaranteed to obtain the optimal value with high probability. $T$ will be polynomial in the problem parameters. More specifically, if each agent has $k$ available actions then the probability of choosing an optimal pair of actions is $\frac{1}{k^2}$. This implies that after $k^3$ trials we get that the optimal joint action will be chosen with probability approaching $1 - e^{-k}$.

## Acknowledgments

We wish to thank the anonymous referees for their useful comments. This work was done while the second author was at the department of computer science, Stanford University. This research was supported by the Israel Science Foundation under grant #91/02-1. The first author is partially supported by the Paul Ivanier Center for Robotics and Production Management. The second author is partially supported by DARPA grant F30602-98-C-0214-P00005.

## Appendix A. The R-max Algorithm

Here we describe the MDP version of the R-max algorithm. R-max can handle zero-sum stochastic games as well, but this capability is not required for the coordination algorithms of this paper. For more details, consult (Brafman & Tennenholtz, 2002).

We consider an MDP $M$ consisting of a set $S = \{s_1, \ldots, s_N\}$ of states in each of which the agent has a set $A = \{a_1, \ldots, a_k\}$ of possible actions. A reward function $R$ assigns an immediate reward for each of the agent's actions at every state. We use $R(s, a)$ to denote the reward obtained by the agent after playing actions $a$ in state $s$. Without loss of generality, we assume all rewards are positive.

In addition, we have a probabilistic transition function, $tr$, such that $tr(s, a, t)$ is the probability of making a transition from state $s$ to state $t$ given that the agent played $a$. It is convenient to think of $tr(s, a, \cdot)$ as a function associated with the action $a$ in the stage-game $s$. This way, all model parameters, both rewards and transitions, are associated with an action at a particular state.

Recall that in our setting, each state of the MDP corresponds to a stage-game and each action of the "agent" corresponds to a joint action of our group of agents.

R-MAX is a model-based algorithm. It maintains its own model $M'$ of the real world model $M$. $M'$ is initialized in an optimistic fashion by making the payoffs for every action in every state maximal. What is special about R-MAX is that the agent always does the best thing according to its current model, i.e., it always exploits. However, if this behavior leads to new information, the model is updated.

Because, initially, the model is inaccurate, exploitation with respect to it may turn out to be exploration with respect to the real model. This is a result of our optimistic initialization of the reward function. This initialization makes state-action pairs that have not been attempted many times look attractive because of the high payoff they yield.

Let $\epsilon > 0$, and let $R_{max}$ denote the maximal possible reward, i.e., $R_{max} \stackrel{\text{def}}{=} \max_{s,a} R(s, a)$. For ease of exposition, we assume that $T$, the $\epsilon$-return mixing time of the optimal policy, is known.

**Initialize:** Construct the following MDP model $M'$ consisting of $N+1$ states, $\{s_0, s_1, \ldots, s_N\}$, and $k$ actions, $\{a_1, \ldots, a_k\}$. Here, $s_1, \ldots, s_N$ correspond to the real states, $\{a_1, \ldots, a_k\}$ correspond to the real actions, and $s_0$ is an additional fictitious state. Initialize $R$ such that $R(s, a) = R_{max}$ for every state $s$ and action $a$. Initialize $tr(s_i, a, s_0) = 1$ for all $i = 0, \ldots, N$ and for all actions $a$ (and therefore, $tr(s_i, a, s_j) = 0$ for all $j \neq 0$).

In addition, maintain the following information for each state-action pair: (1) a boolean value *known/unknown*, initialized to *unknown*; (2) the list of states reached by playing this action in this state, and how many times each state was reached; (3) the reward obtained when playing this action in this state. Items 2 and 3 are initially empty.

**Repeat:**

**Compute and Act:** Compute an optimal $T$-step policy for the current state, and execute it for $T$-steps or until some state-action pair changes its value to *known*.

**Observe and update:** Following each action do as follows: Let $a$ be the action you performed and $s$ the state you were in;

- If the action $a$ is performed for the first time in $s$, update the reward associated with $(s, a)$, as observed.
- Update the set of states reached from $s$ by playing $a$.
- If at this point your record of states reached from this entry contains $K_1 = \max((\lceil \frac{4NTR_{max}}{\epsilon} \rceil^3 \rceil, \lceil -6ln^3(\frac{\delta}{6Nk^2}) \rceil) + 1$ elements, mark this state-action pair as *known*, and update the transition probabilities for this entry according to the observed frequencies.

The basic property of R-MAX is captured by the following theorem:

**Theorem 2** *Let $M$ be an MDP with $N$ states and $k$ actions. Let $0 < \delta < 1$, and $\epsilon > 0$ be constants. Denote the policies for $M$ whose $\epsilon$-return mixing time is $T$ by $\Pi(\epsilon, T)$, and denote the optimal expected return achievable by such policies by $Opt(\Pi(\epsilon, T))$. Then, with* *probability of no less than $1 - \delta$ the* R-MAX *algorithm will attain an expected return of* $Opt(\Pi(\epsilon, T)) - 2\epsilon$ *within a number of steps polynomial in* $N, k, T, \frac{1}{\epsilon},$ *and* $\frac{1}{\delta}$.

Recall that we rely heavily on the fact that R-MAX is deterministic for MDPs. This follows from the existence of a deterministic algorithm for generating an optimal $T$-step policy. Algorithms for generating an optimal $T$-step policy for an MDP are deterministic, except for the fact that at certain choice points, a number of actions may have the same value, and each choice is acceptable. In that case, one needs only generate an arbitrary ordering over the actions and always choose, e.g., the first action. This is why in our algorithm the generation of a shared ordering over the actions plays an important role. Note that, without loss of generality, for MDPs we can restrict our attention to deterministic $T$-step policies. This fact is important in our case because our coordination algorithm would not work if mixed policies were required.